\documentclass[sigconf]{acmart}

\AtBeginDocument{%
  \providecommand\BibTeX{{%
    \normalfont B\kern-0.5em{\scshape i\kern-0.25em b}\kern-0.8em\TeX}}}

\usepackage{epsfig}
\usepackage{graphicx}
\usepackage{amsmath}
\DeclareMathOperator*{\argmax}{arg\,max} 

\usepackage[utf8]{inputenc}
\usepackage[english]{babel}
\usepackage{amsthm}
\newtheorem*{remark}{Remark}
\usepackage{graphics}
\usepackage{booktabs}
\usepackage{makecell}
\usepackage{tabularx}
\usepackage{xcolor}
\usepackage[multiuser,draft,inline,nomargin,marginclue]{fixme}
\fxusetheme{colorsig}
\FXRegisterAuthor{ctc}{actc}{CTC}
\FXRegisterAuthor{hw}{ahw}{HW}
\usepackage{algorithm}
\usepackage[noend]{algpseudocode}
\usepackage{graphicx}
\graphicspath{ {./figures/} }
\usepackage{adjustbox}
\newtheorem{theorem}{Theorem}[section]
\newtheorem{corollary}{Corollary}[theorem]

\usepackage{balance}
\usepackage{subcaption}
\usepackage{cleveref}
\usepackage{hyperref}
\hypersetup{
    colorlinks=true,
    linkcolor=blue,
    filecolor=magenta,      
    urlcolor=blue,
    citecolor=green,
    anchorcolor=green,
}

\newcommand\norm[1]{\left\lVert#1\right\rVert}
\usepackage{booktabs}
\copyrightyear{2020}
\acmYear{2020}
\setcopyright{acmcopyright}
\acmConference[SPAI '20]{Proceedings of the 1st Security
and Privacy on Artificial Intelligent Workshop}{October 6, 2020}{Taipei, Taiwan}
\acmBooktitle{Proceedings of the 1st Security and Privacy on Artificial Intelligent
Workshop (SPAI '20), October 6, 2020, Taipei, Taiwan}
\acmPrice{15.00}
\acmDOI{10.1145/3385003.3410922}
\acmISBN{978-1-4503-7611-2/20/10}
\begin{document}

\fancyhead{}
\title{Toward Few-step Adversarial Training \\ from a Frequency Perspective}

\author{Hans Shih-Han Wang, Cory Cornelius, Brandon Edwards, Jason Martin}
\email{{hans.wang, cory.cornelius, brandon.edwards, jason.martin}@intel.com} 
\affiliation{%
    \institution{Security and Privacy Research, Intel Labs}
    \city{Hillsboro}
    \state{Oregon}
    \country{USA}
    }

    
    
    
\begin{abstract}
We investigate adversarial-sample generation methods from a frequency domain perspective and extend standard $l_{\infty}$ Projected Gradient Descent (PGD) to the frequency domain.
The resulting method, which we call Spectral Projected Gradient Descent (SPGD), has better success rate compared to PGD during early steps of the method.
Adversarially training models using SPGD achieves greater adversarial accuracy compared to PGD when holding the number of attack steps constant.
The use of SPGD can, therefore, reduce the overhead of adversarial training when utilizing adversarial generation with a smaller number of steps.
However, we also prove that SPGD is equivalent to a variant of the PGD ordinarily used for the $l_{\infty}$ threat model. 
This PGD variant omits the sign function which is ordinarily applied to the gradient.
SPGD can, therefore, be performed without explicitly transforming into the frequency domain.
Finally, we visualize the perturbations SPGD generates and find they use both high and low-frequency components, which suggests that removing either high-frequency components or low-frequency components is not an effective defense.
\end{abstract}
\maketitle

\section{Introduction}
An adversarial example is the addition of a perturbation to image such that an image classifier, which correctly classifies the original image, will misclassify this perturbed image.
These adversarial examples often look similar to the original images and may be a concern for safety-critical systems that use these image classifiers.
Many machine learning models are susceptible to these adversarial examples~\cite{szegedy2013intriguing, biggio20_wildpatterns}.

Recent research examined adversarial perturbations in the frequency domain to understand the effectiveness of defenses that utilize this frequency information to mitigate adversarial examples~\cite{Yin2019AFP}.
Others found that filtering the high-frequency components of a perturbation can mitigate adversarial examples~\cite{zhang2019adversarial}.
However, most procedures to generate adversarial examples manipulate individual pixels via gradient updates and not frequency domain information.
We outline these approaches in \Cref{sec:background}.

An alternative approach to generating adversarial examples is to manipulate the individual frequency components of the perturbation.
We explored this approach and extended the $l_{\infty}$ bounded Projected Gradient Descent (PGD) to the frequency domain by explicitly optimizing the frequency components of the perturbation.
We call the new method, Spectral Projected Gradient Descent (SPGD), and formally describe it in \Cref{sec:spgd_attack}.

We also explore how this method can confer robustness to models via adversarial training.
We will show that using SPGD during adversarial training provides competitive robustness while potentially requiring fewer attack steps.
Because adversarial training is computationally expensive owing to the attack procedure used, it is beneficial to reduce the number of attack steps used.
Given a small budget of attack steps, we show in \Cref{sec:experiments} that the SPGD-trained models provide better robustness than the PGD-trained model.
However, we will also prove that SPGD is equivalent to a variant of PGD and show it is not necessary to explicitly optimize in the frequency domain.
We theoretically and empirically show in \Cref{sec:experiments,app:proof} this equivalence for a variant of PGD where the $\textit{sign}$ operation is \emph{not} applied to the gradient.

\section{Background}
\label{sec:background}
In this section, we describe the foundations of our method, including the frequency transformation, gradient-based adversarial generation procedures and adversarial training to produce robust models. 

\subsection{Images in the Frequency Domain}
Images are typically represented as pixel intensity values between 0 and 255 usually with multiple channels representing different colors.
It is also possible to represent each channel of the image as components of a set of frequencies.
One way to do convert pixel intensities to said representation is using the Discrete Cosine Transform (DCT).
The DCT is an orthogonal linear mapping from the pixel domain to the frequency domain, DCT $\colon \mathbf{x} \in \mathbb{R}^{N} \mapsto \mathbf{z} \in \mathbb{R}^{N}$, and IDCT is its inverse~\cite{strang1999discrete, raid2014jpeg}.
Given a natural image $\mathbf{x}$, DCT transforms the image into a representation in the frequency domain $\mathbf{z} = \text{DCT}(\mathbf{x})$ and IDCT transforms this representation back into the pixel domain $\mathbf{x} = \text{IDCT}(\mathbf{z})$.
We chose the DCT to transform perturbations into the frequency domain because it reduces distortions near the image boundary~\cite{raid2014jpeg} better than the discrete Fourier transform does~\cite{strang1999discrete}.

\subsection{Finding Adversarial Examples in the Pixel Domain}
\label{sec:adv_attack1}
Adversarial-example-generation methods seek to cause a model to predict something other than the ground-truth label for a given image by adding a chosen perturbation to the image.
We call this combination of the input image and perturbation an adversarial example when it fools the model.
Conceivably, an adversary could replace the entire image with their desired result by choosing a large perturbation.
However, we study adversaries in a more limited sense.
That is, adversaries can only modify pixels such that the distance between the perturbed image and the input image is small.
For small perturbations, nothing semantic about the image ought to have changed, and therefore the model still ought to correctly classify the image.
To find such perturbations, one can use a variety of existing methods.

As a proxy for the condition that an adversarial example looks similar to the original input, we choose a distance $\epsilon$ and restrict adversarial examples derived from an original input image $x$ to that distance.
More formally, we restrict them to the set $\mathcal{S}(x) = B_{p}(x,\epsilon) \cap \mathcal{I}$ where $B_{p}(x,\epsilon) = \{ x^{\prime} \colon \| x^{\prime} - x\|_{p} \leq \epsilon\}$ is the $l_{p}$ ball centered at $\mathbf{x}$ with radius $\epsilon$ and $\mathcal{I}=\{x \in \mathbb{Z}^{N}: x_i \in [0,255] \ \forall \ i=1,...,N \}$ is the set of valid images, where $N$ is the number of pixels in the image, and $\mathop{\mathbb{Z}}^N$ is the N-fold Cartesian product of copies of $\mathop{\mathbb{Z}}$.
We interpret $\epsilon$ as the hypothetical limitation of the attacker's strength.
As $\epsilon$ increases, the attacker can change more of the input, and, at the limit, they can entirely replace the input.
While $\epsilon$ sets a limit on the distance, we also need to choose a distance metric $l_{p}$.
We focus on $l_{\infty}$ metric for ease of comparison with \citet{madry2017towards}, however, our method could be extended to other threat models.

Fast Gradient Sign Method (FGSM)~\cite{goodfellow2014explaining} is a single step attack that uses the sign of the gradient of the loss function $J$ with respect to the example $\mathbf{x}$: $x^{\prime} = \mathbf{x} + \alpha \cdot sign(\nabla_{x} J(f_{\theta}(x), y; \theta) )$, where $\alpha$ is chosen so that $x^{\prime} \in S(x)$ and $sign(\cdot)$ is component-wise application of $x \rightarrow \frac{x}{|x|}$.
While quick to generate, these perturbations are often ineffective due to gradient masking~\cite{papernot2016distillation, goodfellow2018gradient}.

Project Gradient Descent (PGD)~\cite{madry2017towards} is a multiple-step variant of FGSM. 
To overcome some forms of gradient masking, PGD first randomly initializes the perturbation by uniformly choosing a perturbation within the pre-defined $\epsilon$-ball of the input image.
PGD adds (as opposed to subtracts, because we seek to maximize the model loss) the sign of the gradient (of the loss with respect to the input image) multiplied by some step size $\alpha$ at the attack step $R$.
PGD limits the perturbation by clipping it to the value of $\epsilon$. 
We ought to repeat this process until loss is maximized, but in practice we limit the number of iterations for computational reasons.
More formally, PGD finds an adversarial example:
\begin{equation}
\label{eqn:pgd}
    x_{k+1}^{\prime} = \textit{Proj}_{\mathcal{S}(x)} \left[x_{k}^{\prime} + \alpha \cdot sign(\nabla_{x} J(f_{\theta}(x^{\prime}_{k}), y; \theta)) \right]
\end{equation}
where
\begin{equation*}
\textit{Proj}_{\mathcal{S}(x)}(x^{\prime}) = \underset{[0,255]}{\text{clip}}(x + \underset{[-\epsilon, \epsilon]}{\text{clip}}( \norm{x - x^{\prime}}_{\infty}) )
\end{equation*}
and $\underset{[a, b]}{\text{clip}} (\cdot)$ is component-wise application of:
\begin{equation*}
   x \rightarrow 
  \begin{cases}
      a & \text{if $x < a$} \\
      x & \text{if $a \leq x \leq b$} \\
      b & \text{if $x > b$} \\
  \end{cases}
\end{equation*}

Given an adversarial-example generation method, we perturb each batch of images to create adversarial examples.
As described in \citet{madry2017towards}, PGD is the ``ultimate'' first-order adversary, meaning that other attacks do not find a better maximum so long as the attack only uses first-order gradients. 
Although PGD is a strong attack, it can take many attack steps to converge and this is often the bottleneck in adversarial training.

\subsection{Adversarial Training}
Given a training dataset $\mathcal{D}$ and a classification model $f_{\theta}$ with parameters $\theta$, for each input and ground-truth label $(\mathbf{x}, y) \in \mathcal{D}$ the prediction of $\mathbf{x}$ is $\tilde{y}(x) = \underset{i}{\argmax} f_{\theta}^{i}(\mathbf{x})$, where $i$ is the output index.
The model prediction is successful when the predicted label is equal to the ground-truth label: $\tilde{y}(x) = y$.
In standard training, we seek to solve the following optimization problem:
\begin{equation}
    \label{eqn:train}
    \underset{\theta}{\min} \mathop{\mathbb{E}}_{(x,y)\sim \mathop{\mathbb{D}}} \left[ J(f_{\theta}(x), y) \right]
\end{equation}
$J(x, y)$ can be any loss, but for the purposes of this paper we use the standard cross-entropy loss.
Solving \Cref{eqn:train} yields an $f_{\theta}^{*}$ that classifies inputs from the same distribution as the training set $\mathcal{D}$. 
However, we can use the method in \Cref{sec:adv_attack1}, in particular \Cref{eqn:pgd}, to find adversarial examples that are visually similar to inputs in $\mathcal{D}$ yet cause $f_{\theta}^{*}$ to misclassify them.
To reduce these misclassifications, we seek a robust formulation of the optimization defined in \Cref{eqn:train}.

The most effective method for creating robust models currently is adversarial training~\cite{madry2017towards}.
Adversarial training uses an adversarial-example-generation procedure to generate adversarial examples and incorporates them as training samples on the fly.
The adversarial training process is to find model parameters that minimize the expected loss caused by the adversarial examples that in turn maximize the loss value, $J(x, y;\theta)$ over all allowable perturbations.
More formally, we seek to solve the following optimization problem:
\begin{equation}
    \label{eqn:robust_train}
    \underset{\theta}{\min} \mathop{\mathbb{E}}_{(x,y)\sim \mathop{\mathbb{D}}} \left[ \underset{\delta \in S}{\max} J(f_{\theta}(x + \delta), y) \right].
\end{equation}
\Cref{eqn:robust_train} defines saddle point problem that minimizes the adversarial risk~\cite{madry2017towards, gowal2019alternative}.


\section{Finding Adversarial Examples in the Frequency Domain}
\label{sec:spgd_attack}

\begin{algorithm}[t]
\caption{$\texttt{SPGD-}\alpha\texttt{-R}$ algorithm (untargeted version)}
\label{alg:spgd}
\begin{algorithmic}[1]
\Require Loss function $J(x, y; \theta)$; number of the attack step $R$; clipping range $\epsilon$; natural image $x_{nat}$; momentum factor $\mu$; perturbation $\delta$; step size $\alpha$;
\Procedure{}{}
\State $\delta_{rand} \sim U(-\epsilon, \epsilon)$
\State $x^{\prime}_{0} = x_{nat} + \delta_{rand}$
\State $z^{\prime}_{0}=DCT(x^{\prime}_{0})$
\State $\delta_0 = \nabla_{z}J(z_{0}^{\prime})$

\For{$i = $ 1 to $R$}
    \State $\delta_i = \nabla_{z}J(z_{i-1}^{\prime})$
    \State $\delta_{i} = \delta_{i-1}\mu + \delta_{i}(1-\mu)$ \label{lne:momentum_strategy}

    \State $z_{i}^{\prime} = z_{i-1}^{\prime} + \alpha \delta_{i}$
    \State $x_{i}^{\prime} = IDCT(z_{i}^{\prime})$
    \State  $x_{i}^{\prime} = \text{Proj}_{\mathcal{S}(x)} (x_{i}^{
    \prime})$
    \State $z_{i}^{\prime} = DCT(x_{i}^{\prime})$
\EndFor
\State \Return $x_{R}^{\prime}$ 
\EndProcedure
\end{algorithmic}
\end{algorithm}

To find adversarial examples in the frequency domain, we modify PGD to optimize directly in the frequency domain.
Our perturbation generation procedure, Spectral Projected Gradient Descent (SPGD), uses a frequency transformations--DCT and IDCT--to perform gradient updates in the frequency domain. 
SPGD converts the randomly initialized examples into the DCT domain (frequency domain), which is represented as DCT coefficients.
It then computes the gradient of the loss with respect to the frequency variables and adds a scaled version of this to the DCT coefficients of the input.
To make sure the perturbed input is in the allowable set of inputs, SPGD uses IDCT to convert this result back to the input domain and then projects the input to the allowable set of inputs.
This procedure is formally outlined in \Cref{alg:spgd}.

\subsection{Selecting Parameters}
We arrived at the large step sizes used for SPGD in our experiments based on which step size caused lower adversarial accuracy for the CIFAR-10 dataset when acting on the adversarially trained model introduced by \citet{madry2017towards}. We performed a preliminary search for the proper step size, $\alpha$, indicating the starting point of 10,000, and then we searched over different scales of the parameter from 10,000 to 1,000,000,000, and found the best empirical value to be 75,000,000. 
This finding may be contingent on the choice of the model structure and optimizer, but such exploration is beyond the scope of this work.
We leave further exploration to future work.

The gradients seen during adversarial example generation varies between iterations causing that direction of optimization to change at each step, thus slowing convergence and perhaps leading to ineffective adversarial examples~\cite{dong2018boosting}.
Momentum accumulates gradients across iterations so that the historic gradient dampens large changes~\cite{dong2018boosting}.
To generate more effective adversarial examples, we use the momentum strategy in \citet{dong2018boosting}.
We uniformly searched over the momentum factor $\mu$ from $0.0$ to $1.0$, and empirically found the best value that minimized adversarial accuracy to be $0.75$.
Using momentum decreases the adversarial accuracy of the same model from $48.49\%$ to $45.68\%$.
We apply the idea of momentum for the gradient update in the frequency domain at the \Cref{lne:momentum_strategy} in \Cref{alg:spgd} to improve the performance of adversarial generation.

\section{Experiments}
\label{sec:experiments}
In this section we study SPGD using the CIFAR-10 dataset.
We also show experiments on MNIST in \Cref{app:mnist}
We performed experiments to demonstrate that SPGD for adversarial training can reduce the number of required attack steps while retaining competitive robustness.
For a fair comparison in our experiments, we augmented PGD with momentum.
However, the best momentum value we found was 0.
In the experiments below, we used a step size of 75,000,000 for SPGD and step size 2 for PGD during adversarial training.

We used the wide ResNet-18 architecture~\cite{he2016deep} for all CIFAR-10 experiments and tested these models against PGD and SPGD adversaries with attack strength $\epsilon=8$.~\footnote{See: \url{https://github.com/MadryLab/cifar10_challenge}}
The natural accuracy of every adversarial-trained model was over 87\%.
We run 10 steps of PGD or SPGD as our adversary during adversarial training, and used 20 iterations of PGD or SPGD as our adversary for evaluation. 
In order to understand the impact of reducing the number of attack steps used during training, we reduce these steps from 10 to 5, holding all other training and testing parameters constant.

\subsection{SPGD Requires Fewer Attack Steps Than PGD}
\begin{table*}[t]
  \caption{We evaluate the PGD-adversarially-trained model with the PGD or SPGD attacks. Adversarial loss values and adversarial accuracy change over attack steps. The PGD attack increases adversarial loss faster when the PGD attack step increases from 1 to 8. The SPGD attack increases adversarial loss faster than both cases of PGD attack and provides earlier attack convergence.  }
  \begin{adjustbox}{width=\textwidth,center}
    \label{tab:adv_attack1}
    \begin{tabular}{l*{20}c}
      \toprule
      \multicolumn{21}{c}{CIFAR-10} \\
      \multicolumn{1}{r}{Attack Step (R)} & 1 & 2 & 3 & 4 & 5 & 6 & 7 & 8 & 9 & 10 & 11 & 12 & 13 & 14 & 15 & 16 & 17 & 18 & 19 & 20\\
      \midrule
      Attack Method & \multicolumn{20}{l}{Adversarial Accuracy} \\
      \cmidrule(r){1-1} \cmidrule(l){2-21}
      PGD-2-R & 79.95\% &    72.37\% &    65.33\% &    59.98\% &    55.53\% &    52.17\% &    49.65\% &    48.47\% &    47.88\% &    47.31\% &    46.89\% &    46.64\% &    46.40\% &    46.43\% &    46.30\% &    46.20\% &    46.07\% &    46.03\% &    45.96\% &    45.88\% \\
      PGD-8-R & 62.71\% &    51.78\% &    48.43\% &    47.23\% &    46.63\% &    46.49\% &    46.30\% &    46.25\% &    46.21\% &    46.11\% &    46.14\% &    46.11\% &    46.02\% &    46.00\% &    46.01\% &    45.97\% &    45.92\% &    45.92\% &    45.95\% &    45.94\%\\
      SPGD-75e6-R & 56.77\% &    50.16\% &    48.18\% &    47.02\% &    46.49\% &    46.31\% &    46.20\% &    46.07\% &    46.05\% &    45.96\% & 45.89\% &    45.86\% &    45.85\% & 45.82\% & 45.81\% &    45.72\% &    45.72\% &    45.74\% &    45.69\% &    45.67\%\\
      \addlinespace[4pt]
      Attack Method & \multicolumn{20}{l}{Adversarial Loss}\\
      \cmidrule(r){1-1} \cmidrule(l){2-21}
      PGD-2-R & 0.7699&    1.1544&    1.5710&    1.9754 &    2.3421&    2.6377&    2.8567&    2.9788 &    3.0438&    3.0965 & 3.1331 & 3.1628 &    3.1836 &    3.2007 &    3.2111 &    3.2212 &    3.2276 &    3.2339 &    3.2395&    3.2430\\
      PGD-8-R & 1.7476 &    2.6415 &    2.9483 &    3.0501 &    3.0888 &    3.1041 &    3.1128 &    3.1181 &    3.1202 &    3.1221 &    3.1224 &    3.1265 &    3.1276 &    3.1267 &    3.1295 &    3.1289 &    3.1303 &    3.1297&    3.1297&    3.1331\\
      SPGD-75e6-R & 2.2452&    2.7091 &    2.9529 &    3.0695 &    3.1239 &    3.1541 & 3.1723 &    3.1847 &    3.1945 &    3.2014 &    3.2068 &    3.2115 &    3.2156&    3.2185 &    3.2215 &    3.2235 &    3.2263 &    3.2281 &    3.2300 &    3.2310 \\
      \bottomrule
    \end{tabular}
  \end{adjustbox}
\end{table*}

As shown in \Cref{tab:adv_attack1}, we used either the SPGD method or the PGD method to attack the state-of-the-art PGD-trained model, evaluating the effect of the PGD and SPGD attacks using two indicators, adversarial accuracy, and adversarial loss.
At early attack steps, SPGD is stronger than PGD (indicated by lower adversarial accuracy and higher adversarial loss), though, at their final steps, the two attacks are on par with each other according to both indicators.
In use cases for which the computational overhead of adversarial training is a concern, the use of SPGD with fewer steps may be a good choice to consider.

\subsection{Adversarial Training using SPGD}
\begin{table}[ht]
    \caption{The adversarial accuracy of the adversarially-trained models with SPGD or PGD on the CIFAR10 dataset. We adversarially trained the models with $\epsilon=8$ by using either the SPGD attack or the PGD attack. We also reduce the number of attack steps used during adversarial training from 10 to 5. We used either the SPGD or PGD attack to evaluate the adversarially-trained models and found that the SPGD-trained model retained better robustness when using 5 attack steps during training than the corresponding PGD-trained model.}
    \label{tab:security_curve1}
    \begin{adjustbox}{width=\columnwidth}
        \begin{tabular}{lccccccccc}
            \toprule
            \multicolumn{10}{c}{CIFAR-10} \\
            \multicolumn{2}{r}{Attack Strength ($\epsilon$)} & 0 & 1 & 2 & 4 & 8 & 16 & 32 & 64\\
            \midrule
            Attack Method & Defense Model & \multicolumn{7}{l}{Adversarial Accuracy} \\
            \cmidrule(r){1-1} \cmidrule(lr){2-2} \cmidrule(l){3-10}
            PGD-2-20 & SPGD-75e6-10 & 87.63\% & 84.63\% & 80.67\% & 71.62\% & 49.54\% & 18.38\% &  4.36\% & 0.34\% \\
            PGD-2-20 & PGD-2-10  & 87.25\% & 83.59\% & 79.07\% & 68.50\% & 47.26\% & 28.92\% & 18.37\% & 5.07\% \\
            PGD-2-20 & SPGD-75e6-5  & 87.76\% & 84.64\% & 80.75\% & 70.83\% & 48.85\% & 17.32\% &  3.62\% & 0.38\% \\
            PGD-2-20 & PGD-2-5   & 88.47\% & 84.46\% & 79.83\% & 69.24\% & 44.95\% & 18.39\% &  7.41\% & 0.75\% \\
            \addlinespace[4pt]
            SPGD-75e6-20 & SPGD-75e6-10 & 87.63\% & 84.62\% & 80.68\% & 71.69\% & 49.60\% & 15.08\% & 0.40\% & 0.00\% \\
            SPGD-75e6-20 & PGD-2-10  & 87.25\% & 83.59\% & 79.09\% & 68.45\% & 45.80\% & 15.15\% & 1.30\% & 0.02\% \\
            SPGD-75e6-20 & SPGD-75e6-5  & 87.76\% & 84.64\% & 80.74\% & 70.95\% & 49.12\% & 14.08\% & 0.28\% & 0.00\% \\
            SPGD-75e6-20 & PGD-2-5   & 88.47\% & 84.46\% & 79.85\% & 69.26\% & 44.74\% & 15.10\% & 2.22\% & 0.08\% \\
            \bottomrule
        \end{tabular}
    \end{adjustbox}
\end{table}

We found that SPGD-trained models provide competitive robustness to the PGD-trained models.
As shown in \Cref{tab:security_curve1}, the SPGD-75e6-10-trained model is more robust than the PGD-2-10-trained model at $\epsilon=8$.
When attacking with PGD at $\epsilon=8$, the SPGD-75e6-10-trained model has an accuracy of 49.54\%, which is better than the PGD-2-10-trained model's accuracy of 47.26\% by 2.28 percentage points.
When attacking with SPGD at $\epsilon=8$, the SPGD-75e6-10-trained model has accuracy of 49.60\%, which is better than the PGD-2-10-trained model's accuracy of 45.80\% by 3.8\% points.

As shown in \Cref{tab:security_curve1}, the adversarial accuracy of the SPGD-75e6-5-trained model shows a smaller performance drop from the SPGD-75e6-10-trained model.
The SPGD attack requires fewer attack steps to achieve higher adversarial loss than the PGD, which helps improve the efficiency of adversarial training by reducing the number of attack steps.

When attacked by PGD with $\epsilon=8$, the SPGD-75e6-5-trained model performs at 48.85\% adversarial accuracy, which is lower than the SPGD-75e-10-trained model performance of 49.54\% by only 0.69\% points. In contrast, the PGD-2-5-trained model performs at 44.95\% adversarial accuracy, which is lower than the PGD-2-10-trained model's performance of 47.26\% by 2.31\% points.
When attacked by SPGD with $\epsilon=8$, the SPGD-75e6-5-trained model performs at 49.12\% adversarial accuracy, which is lower than the SPGD-75e6-10-trained model's performance of 49.60\% by only 0.48\% points. In contrast, the PGD-2-5-trained model performs at 44.74\% adversarial accuracy, which is lower than the PGD-2-10-trained model's performance of 45.80\% by 1.06\% points.
Whether being attacked by SPGD or PGD, the model trained with more attack steps retains robustness much better than the model trained with less attack steps.

When attacked by PGD with $\epsilon=8$, the SPGD-75e6-5-trained model performs at 48.85\% adversarial accuracy, which is higher than the PGD-2-5-trained model's performance of 44.95\% by 3.90\% points.
When attacked by SPGD with $\epsilon=8$, the SPGD-75e6-5-trained model performs at 49.12\% adversarial accuracy, which is higher than the PGD-2-5-trained model's performance of 44.74\% by 4.38\% points.
Whether being attacked by SPGD or PGD, the SPGD-trained model retains robustness much better than the PGD-trained model by using fewer attack steps during adversarial training..

\section{Discussion}
We have demonstrated the results of adversarial training with SPGD, and we want to discuss how SPGD perturbation is compared with PGD perturbation qualitatively from frequency domain and pixel domain perspectives.

\subsection{Visualizing SPGD and PGD Perturbations}
\begin{figure*}[htp]
  \begin{subfigure}[t]{0.33\textwidth}
    \includegraphics[width=\textwidth]{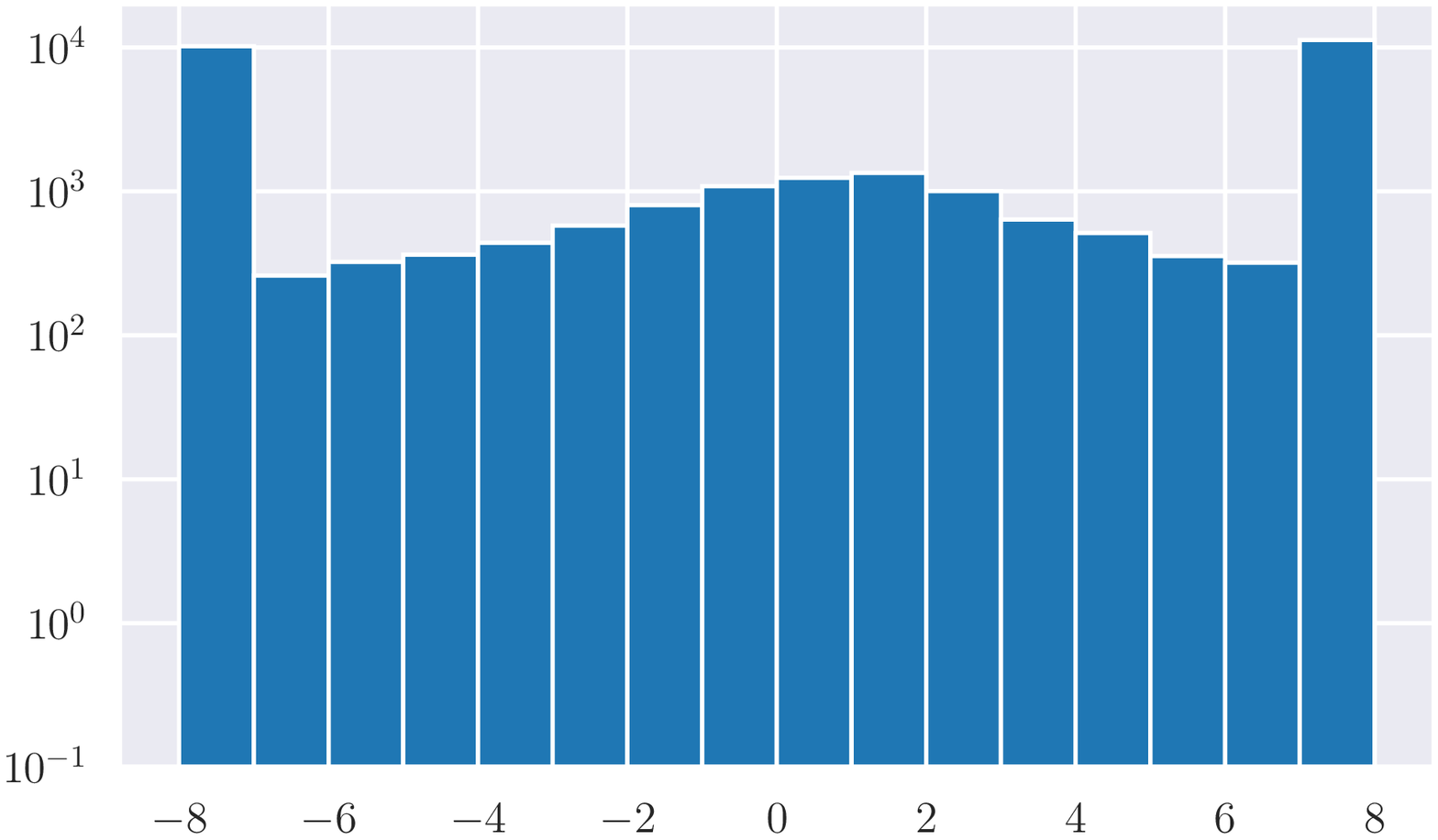}
    \caption{Histogram of SPGD-75e6-1}
    \label{fig:spgd2_1img}
  \end{subfigure}
  \hfill
  \begin{subfigure}[t]{0.33\textwidth}
    \includegraphics[width=\textwidth]{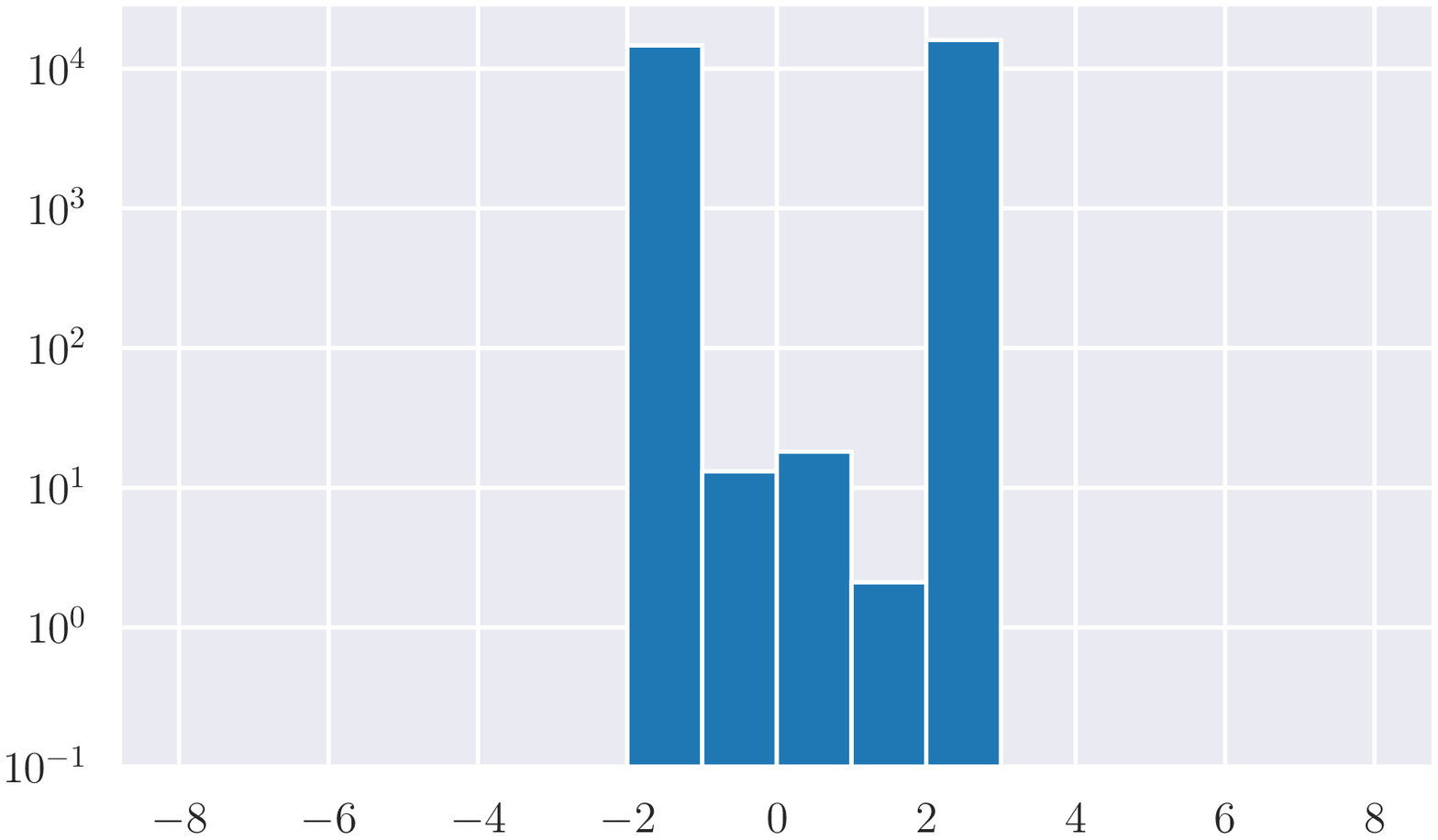}
    \caption{Histogram of PGD-2-1}
    \label{fig:pgd2_1img}
  \end{subfigure}
  \hfill
  \begin{subfigure}[t]{0.33\textwidth}
    \includegraphics[width=\textwidth]{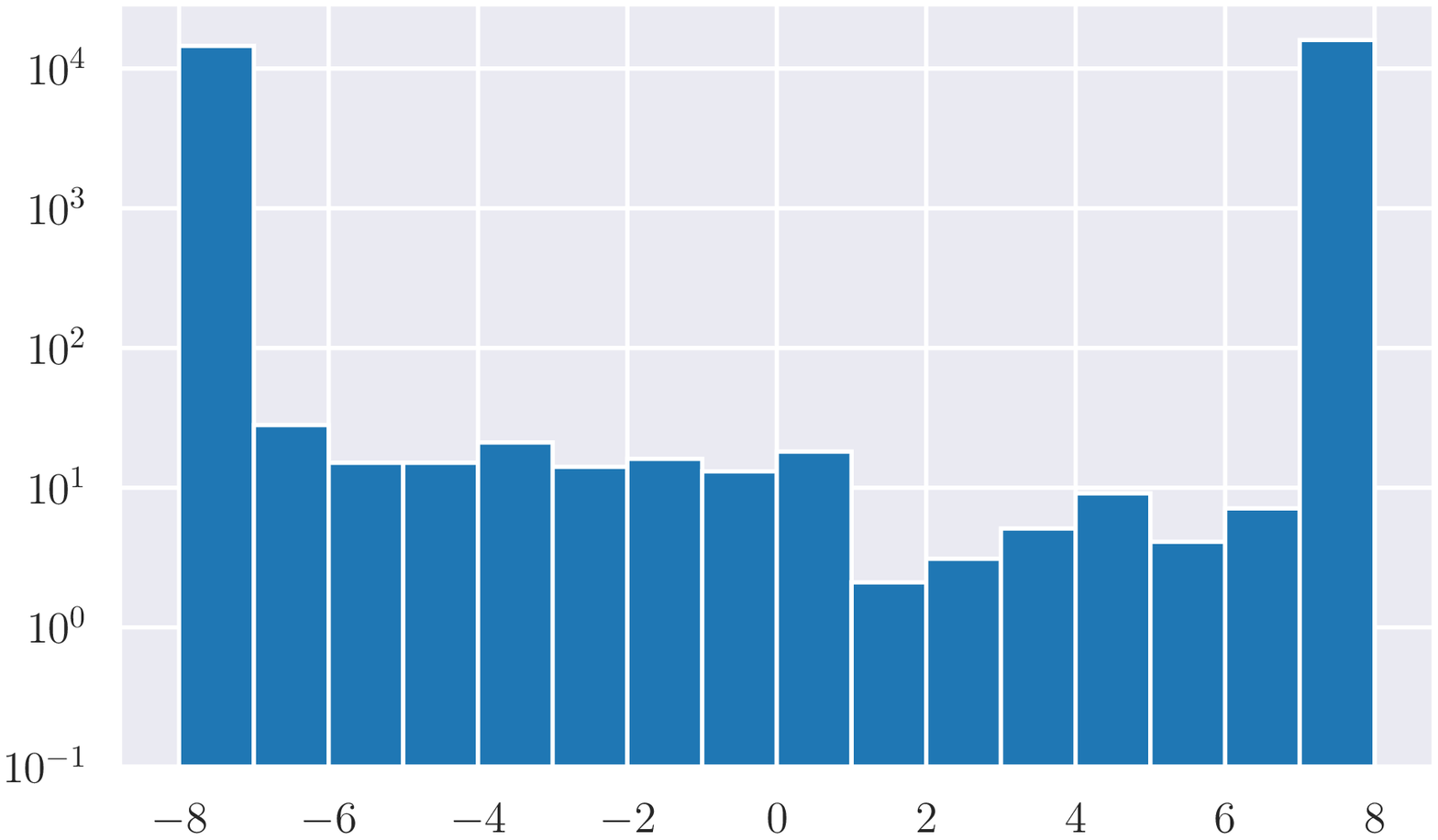}
    \caption{Histogram of PGD-8-1}
    \label{fig:pgd8_1img}
  \end{subfigure}
  \caption{
    Histograms of component values for the specified perturbation generation procedure at the first step.
    The perturbations components of PGD shown in \Cref{fig:pgd2_1img,fig:pgd8_1img} are concentrated near the boundary of the step size while the small components of SPGD (\Cref{fig:spgd2_1img}) are more evenly distributed.
    We show these plots over multiple steps of each perturbation generation procedure in \Cref{app:distributions}.
  }
\end{figure*}

To understand why SPGD requires fewer attack steps during adversarial training, we randomly sampled 10 images from the CIFAR-10 dataset and calculated the distribution of the components of the perturbations after the first step as shown in \Cref{fig:spgd2_1img}.
\Cref{fig:pgd2_1img} shows the components of a PGD-generated perturbation at the first step are constrained to value bins nearer to zero than for SPGD due to the step size, so it takes more attack steps to achieve a similar profile as SPGD. 
Note that the final profiles have dominant value mass at the epsilon boundary shown in Appendix B. 
Even when we used a PGD step size as large as epsilon, the early stage of the PGD attack requires more steps than SPGD to drive down the adversarial accuracy, as shown in \Cref{tab:adv_attack1}. 

In \Cref{fig:spgd2_1img,fig:pgd8_1img}, we see that SPGD produces more perturbation components that are less than $\epsilon$ than PGD does. 
We conjecture this is the key difference between SPGD and PGD.
The SPGD attack allows for the extremely small gradient values to not be weighted as heavily--with more of them resulting in perturbation components that are less than epsilon. This in turn allows for a higher degree of freedom in movement during the optimization process. 

\begin{figure}[htb]
    \centering
    \includegraphics[scale=0.27]{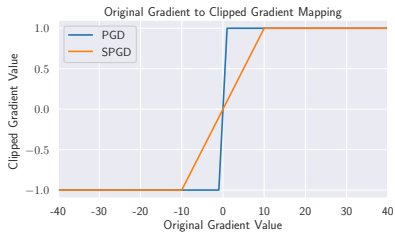}
    \caption{The plot shows the value mapping of perturbation components for PGD and SPGD at the first attack step. The mapping of sign operation and $\epsilon$-ball projection indicated by the blue color quantizes all values into 1, -1 or 0. The mapping of only $\epsilon$-ball projection indicated by the orange color has a linear piece-wise function within the $\epsilon$-ball.   }
    \label{fig:grad_value_mapping}
\end{figure}

To further explore this difference in weighting of gradient values, in \Cref{fig:grad_value_mapping}, we illustrate the difference during the first attack step between the sign function utilized in PGD and what can be considered as a replacement for the sign function when performing SPGD with large step size. 
The $l_{\infty}$ attack constraint allows the attacker to apply a perturbation whose components have value in $[-\epsilon,\epsilon]$. The sign of a given gradient component indicates the change in the loss, local to the point where the gradient is being calculated. In the case of a linear loss function, this local change is sustained globally, so that $\epsilon \cdot sign(\nabla_{x}f)$ will achieve the optimal loss given the constraints. In the case where the relationship between model parameters and loss is not linear; however, this perturbation is good only in as much as the local change in loss is sustained globally. 
Although the SPGD attack is sub-optimal if all gradient components predict global growth well, in the case of the neural network, when applied with a large step size, its updates follow a heuristic wherein the smaller gradient components are considered more likely to change sign as we step away from where we are computing the gradient during optimization. Therefore the perturbation component should be smaller than $\epsilon$ when the corresponding gradient component is small relative to some threshold.
On the other hand, the PGD attack applies the sign function to the gradient, treating all gradient values (even extremely small ones) as equally predictive of global trends. 
All perturbation components will have the same $\epsilon$ in absolute value,  as shown in \Cref{fig:grad_value_mapping}. 

\begin{table*}[t]
  \caption{Adversarial accuracy and adversarial loss resulting from SPGD-75e6-R and its pixel-domain version, NoSignPGD-75e6-R (shown to be theoretically equivalent in the \Cref{app:proof}) acting on the PGD adversarially trained model introduced by \citet{madry2017towards} for various values of attack step. The results are consistent with the theoretical equivalence (accounting for floating point errors).}
  \label{tab:adv_attack_spgd_pgd_same}
  \begin{adjustbox}{width=\textwidth,center}
  \label{adv_attack2}
      \begin{tabular}{l*{19}cc}
        \toprule
        \multicolumn{21}{c}{CIFAR-10} \\
        Attack Step (R) & 1 & 2 & 3 & 4 & 5 & 6 & 7 & 8 & 9 & 10 & 11 & 12 & 13 & 14 & 15 & 16 & 17 & 18 & 19 & 20\\
        \midrule
        Attack & \multicolumn{20}{l}{Adversarial Accuracy of \citet{madry2017towards}} \\
        \cmidrule(r){1-1} \cmidrule(l){2-21}
        SPGD-75e6-R        & 56.77\% & 50.16\% & 48.92\% & 47.02\% & 46.49\% & 46.31\% & 46.2 \%& 46.07\% & 46.05\% & 45.96\% & 45.89\% & 45.86\% & 45.85\% & 45.82\% & 45.81\% & 45.72\% & 45.72\% & 45.74\% & 45.69\% & 45.67\% \\
        NoSignPGD-75e6-R & 56.76\% & 50.20\% & 48.81\% & 47.06\% & 46.52\% & 46.28\% & 46.24\% & 46.13\% & 45.95\% & 45.95\% & 45.87\% & 45.88\% & 45.89\% & 45.82\% & 45.78\% & 45.82\% & 45.81\% & 45.74\% & 45.70\% & 45.66\% \\
        \addlinespace[4pt]
        Attack & \multicolumn{20}{l}{Adversarial Loss of \citet{madry2017towards}} \\
        \cmidrule(r){1-1} \cmidrule(l){2-21}
        SPGD-75e6-R        & 2.2449 & 2.7091 & 2.7350 & 3.0695 & 3.1240 & 3.1541 & 3.1723 & 3.1847 & 3.1945 & 3.2014 & 3.2068 & 3.2115 & 3.2156 & 3.2185 & 3.2215 & 3.2235 & 3.2263 & 3.2281 & 3.2300 & 3.2310 \\
        NoSignPGD-75e6-R & 2.2453 & 2.7061 & 2.7348 & 3.0648 & 3.1217 & 3.1524 & 3.1698 & 3.1840 & 3.1945 & 3.2012 & 3.2060 & 3.2126 & 3.2175 & 3.2190 & 3.2211 & 3.2219 & 3.2246 & 3.2262 & 3.2286 & 3.2300 \\
        \bottomrule
        \end{tabular}
  \end{adjustbox}
\end{table*}

The attack effect of SPGD covers both $l_2$ and $l_\infty$ norm ball after the $l_{\infty}$ projection. 
We prove that SPGD can be computed in the pixel domain is a variant of PGD (omitting application of the sign function), where we set the gradient update to be: $\alpha\cdot\nabla_{x}f$ as shown in \Cref{eqn:SPGD_pixel} (with a proof of this shown in \Cref{app:proof}):
\begin{equation}
    \label{eqn:SPGD_pixel}
    \text{Proj}_{x^{\prime} \in \mathcal{S}(x)} (\alpha \cdot \nabla_{x} J) )
\end{equation}
In \Cref{tab:adv_attack_spgd_pgd_same}, we show that the SPGD and the variant of the PGD algorithm are also empirically close to each other with small amounts of floating-point error.

\subsection{Are Adversarial Perturbations Necessarily High Frequency?}
\begin{figure}
    \centering
    \includegraphics[scale=0.35]{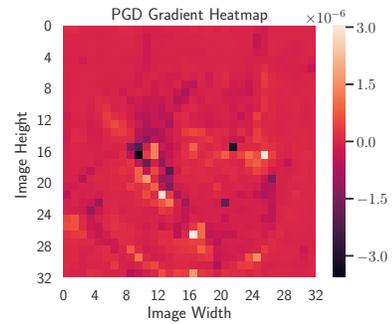}
    \caption{The heat map of the PGD gradient components at the first attack step in the pixel domain indicates the pixels triggered to increase adversarial loss is a certain pattern.}
    \label{fig:gradients_pgd}
\end{figure}

\begin{figure}
    \centering
    \includegraphics[scale=0.35]{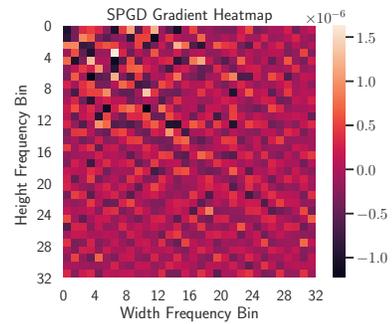}
    \caption{The heat map of the SPGD gradient components at the first attack step in the frequency domain indicates the frequency bins triggered to increase adversarial loss are from both low frequency and high-frequency areas.}
    \label{fig:gradients_spgd}
\end{figure}

Some research found that the adversarial perturbation tends to be high-frequency components, and it might be mitigated by focusing on high frequency signals~\cite{bandlimiting_NN}. 
We plot counterexamples in \Cref{fig:gradients_pgd,fig:gradients_spgd} that adversarial perturbation could exist not only in the high-frequency band but also in the low-frequency band.
We randomly picked one image from CIFAR10 and calculated the raw gradient at the first step by using PGD and SPGD in the pixel domain and frequency domain, respectively. 
In \Cref{fig:gradients_pgd}, the raw gradient of PGD in the pixel domain indicates which part of pixels is triggered for perturbation. 
In \Cref{fig:gradients_spgd}, the SPGD method provides interpretation from the frequency aspect to understand the adversarial examples. 
The raw gradient of SPGD in the frequency domain shows that some frequency bins got triggered to generate the adversarial perturbation, and many of the triggered bins are concentrated in the lower frequency area while some of the triggered bins are concentrated in the higher frequency area. 
As shown, some frequency bins need to be heightened, and some need to be weakened, so the pattern of the grid has both positive and negative value changes. 
Furthermore, we conjecture that the adversarial examples could exist anywhere in the frequency domain regardless of high frequency or low frequency. 
Therefore, simply removing or blurring some frequency components may not solve the root cause of the model susceptibility; instead, it may simply patch some discovered vulnerability ignoring some other vulnerability not being discovered yet.


\section{Conclusion}
We described an extension of PGD called SPGD and demonstrated its effectiveness at attacking models and improving adversarial training.
SPGD finds adversarial examples with higher loss in fewer attack steps than the PGD attack. 
However, we also proved that SPGD is a variant of PGD attack in the pixel domain when the $\textit{sign}$ operation is omitted.
On CIFAR-10, our SPGD-trained Wide ResNet-18 model can achieve higher adversarial accuracy than the PGD-trained model. 
When the adversarial training steps are reduced to a half, our SPGD-trained model is on par with the PGD-trained model and can remain competitive.
That is, we can use fewer attack steps to train a competitive model by using SPGD in adversarial training. 
Lastly, our visualization of these SPGD-generated adversarial perturbations shows their components are not necessarily concentrated at high frequencies but are also found at low frequencies.
In fact, we conjecture the adversarial perturbations can happen everywhere regardless of high frequency or low frequency. 
We hope this paper motivates more exploration in the frequency domain for adversarial analysis and serve as an inspiration for a broader view on the landscape of the adversarial examples. 




{
\small
\bibliographystyle{ACM-Reference-Format}
\balance 
\bibliography{reference}
}
\appendix
\section{Proof of SPGD-PGD Equivalence}
\label{app:proof}
\begin{theorem}
\label{theorem_1}
Let $\mathbf{A} \in \mathbb{R}^{N\times N}$be a matrix corresponding to a transformation $\mathbf{z} \mapsto \mathbf{x}$ with $\mathbf{x}=\mathbf{A} \mathbf{z}$. If $f\colon \mathbb{R}^N \mapsto \mathbb{R}$, then $\nabla_{\mathbf{z}} f = \mathbf{A}^{\intercal} \nabla_{\mathbf{x}} f$
\end{theorem}

\begin{corollary}
    \label{col_1}
    With $\mathbf{A},\ f,\ \mathbf{x}\ \text{and}\ \mathbf{z}$ as in \Cref{theorem_1}, if $\mathbf{A}$ is orthogonal and $\alpha \in \mathbb{R}$, then $\mathbf{A} \cdot \alpha \nabla_{\mathbf{z}}f = \alpha \nabla_{\mathbf{x}}f$
\end{corollary}

\begin{corollary}
    \label{col_2}
      The SPGD algorithm with learning rate $\alpha$ is equivalent to the PGD algorithm with the step size equal to $\alpha$ and no sign function applied to the gradient. That is, the SPGD update is equivalent to the form:\\ 
      \begin{align*}\mathbf{x}_{n+1} = \text{Proj}_{ \mathcal{S}(\mathbf{x})}(\mathbf{x}_{n} + \alpha\nabla_{\mathbf{x}} J))\end{align*}\\ 
      whereas the PGD update is given by:\\ 
      \begin{align*}\mathbf{x}_{n+1} = \text{Proj}_{ \mathcal{S}(\mathbf{x})}(\mathbf{x}_{n} + \alpha \cdot sign(\nabla_{\mathbf{x}}J))\end{align*}\\
      where $\mathbf{x}_1, \mathbf{x}_2, \cdots$ is the sequence of attack iterations, and $J$ is the loss associated to the classifier. 
\end{corollary}

\begin{theorem}[theorem 1]
\label{theorem_1_app}
Let $\mathbf{A} \in \mathbb{R}^{N\times N}$be a matrix corresponding to a transformation $\mathbf{z} \mapsto \mathbf{x}$ with $\mathbf{x}=\mathbf{A} \mathbf{z}$. If $f\colon \mathbb{R}^N \mapsto \mathbb{R}$, then $\nabla_{\mathbf{z}} f = \mathbf{A}^{\intercal} \nabla_{\mathbf{x}} f$
\end{theorem}

\begin{proof}
\[
\mathbf{x} = \mathbf{A} \mathbf{z}\\
\]
\[
\begin{split}
    \begin{pmatrix}
        x_1\\
        \vdots\\
        x_N\\
    \end{pmatrix}
    &= \begin{pmatrix}
        a_{11} & \cdots & a_{1N} \\
        \vdots & \ddots & \vdots \\
        a_{N1} & \cdots & a_{NN} \\
    \end{pmatrix}
    \begin{pmatrix}
        z_1 \\
        \vdots\\
        z_N\\
    \end{pmatrix}\\
    &= \begin{pmatrix}
        \sum_{j=1}^{N} a_{1j}z_{j}\\
        \vdots\\
        \sum_{j=1}^{N} a_{ij}z_{j}\\
        \vdots\\
        \sum_{j=1}^{N} a_{Nj}z_{j}\\
    \end{pmatrix}
\end{split}
\]

\[
\nabla_{z}\mathbf{x}
=
\begin{pmatrix}
    \frac{\partial{x_i}}{\partial z_1},\cdots,\frac{\partial{x_i}}{\partial z_N} 

\end{pmatrix}
\]

\begin{equation}
\text{Jacobian matrix}\ \mathcal{J}_{A}
=
\begin{pmatrix}
    \frac{\partial x_1}{\partial z_1} &\cdots & \frac{\partial x_1}{\partial z_N}\\ 
    \vdots &\ddots &\vdots \\
    \frac{\partial x_N}{\partial z_1} &\cdots &\frac{\partial x_N}{\partial z_N}
\end{pmatrix}
=
\begin{pmatrix}
\mathbf{a}_{i,j}
\end{pmatrix}
=
\mathbf{A}
\end{equation}
\begin{equation}
\begin{split}
    \nabla_{z}f
    &=
    \begin{pmatrix}
        \frac{\partial f}{\partial z_1}\\
        \vdots\\
        \frac{\partial f}{\partial z_N}\\
    \end{pmatrix}\\
    &=
    \begin{pmatrix}
        \sum_{i=1}^{N} \frac{\partial{f}}{\partial{x_i}} \frac{\partial{x_i}}{z_1}\\
        \vdots\\
        \sum_{i=1}^{N} \frac{\partial{f}}{\partial{x_i}} \frac{\partial{x_i}}{z_N}\\
    \end{pmatrix}\\
    &=
    \begin{pmatrix}
        \frac{\partial x_1}{\partial z_1} &\cdots & \frac{\partial x_N}{\partial z_1}\\ 
        \vdots &\ddots &\vdots \\
        \frac{\partial x_1}{\partial z_N} &\cdots &\frac{\partial x_N}{\partial z_N}\\
    \end{pmatrix}
    \begin{pmatrix}
        \frac{\partial f}{\partial x_1}\\
        \vdots\\
        \frac{\partial f}{\partial x_N}\\
    \end{pmatrix}\\
    &=
    \begin{pmatrix}
        a_{11} & \cdots & a_{N1}\\
        \vdots & \ddots & \vdots\\
        a_{1N} & \cdots & a_{NN}\\
    \end{pmatrix}
    \begin{pmatrix}
        \frac{\partial f}{\partial x_1}\\
        \vdots\\
        \frac{\partial f}{\partial x_N}\\
    \end{pmatrix}\\
    &=
    \mathbf{A}^\intercal  \nabla_x f\\
\end{split}
\end{equation}
\end{proof}


\begin{corollary}
    \label{col_1_app}
    With $\mathbf{A},\ f,\ \mathbf{x}\ \text{and}\ \mathbf{z}$ as in \Cref{theorem_1_app}, if $\mathbf{A}$ is orthogonal and $\alpha \in \mathbb{R}$, then $\mathbf{A} \cdot \alpha \nabla_{\mathbf{z}}f = \alpha \nabla_{\mathbf{x}}f$
\end{corollary}

\begin{proof}
\[
    \begin{split}
        \mathbf{A} \cdot \alpha \nabla_{\mathbf{z}}f 
        &= 
        \mathbf{A} \cdot \alpha \mathbf{A}^{\intercal} \nabla_{\mathbf{x}}f\\
        &=
        \alpha \cdot \mathbf{A} \mathbf{A}^{\intercal} \nabla_{\mathbf{x}}f\\
        &=
        \alpha \cdot \mathbf{A} \mathbf{A}^{-1} \nabla_{\mathbf{x}}f\\
        &=
        \alpha \nabla_{\mathbf{x}}f
    \end{split}
\]
\end{proof}

\begin{corollary}
    \label{col_2_app}
      The SPGD algorithm with learning rate $\alpha$ is equivalent to the PGD algorithm with the step size equal to $\alpha$ and no sign function applied to the gradient. That is, the SPGD update is equivalent to the form:\\ 
      \begin{align*}\mathbf{x}_{n+1} = \text{Proj}_{ \mathcal{S}(\mathbf{x})}(\mathbf{x}_{n} + \alpha\nabla_{\mathbf{x}} J))\end{align*}\\ 
      whereas the PGD update is given by:\\ 
      \begin{align*}\mathbf{x}_{n+1} = \text{Proj}_{ \mathcal{S}(\mathbf{x})}(\mathbf{x}_{n} + \alpha \cdot sign(\nabla_{\mathbf{x}}J))\end{align*}\\
      
      where $\mathbf{x}_1, \mathbf{x}_2, \cdots$ is the sequence of attack iterations, and $J$ is the loss associated with the classifier. 

\end{corollary}
\begin{proof}
    Let $\mathbf{A}$ the matrix of the inverse discrete cosine transform. 
    The original SPGD algorithm update is in the frequency domain such that $\mathbf{A}\cdot \alpha\nabla_{\mathbf{z}} J$. 
    According to \Cref{col_1}, we infer to get $\mathbf{A}\cdot \alpha\nabla_{\mathbf{z}} J=\alpha\nabla_{\mathbf{x}} J$.
    The PGD algorithm update is given from ~\cite{goodfellow2014explaining}.

\end{proof}

\begin{remark}
With the \Cref{theorem_1_app}, the \Cref{col_1_app}, and the \Cref{col_2_app}, we show that the SPGD algorithm optimizing in the frequency domain is equivalent to a variant of the PGD algorithm optimizing in the pixel domain. 
\end{remark}

\section{Distributions of Gradient Components}
\label{app:distributions}
\begin{figure*}[tbp]
    \centering
    \includegraphics[width=0.95\textwidth]{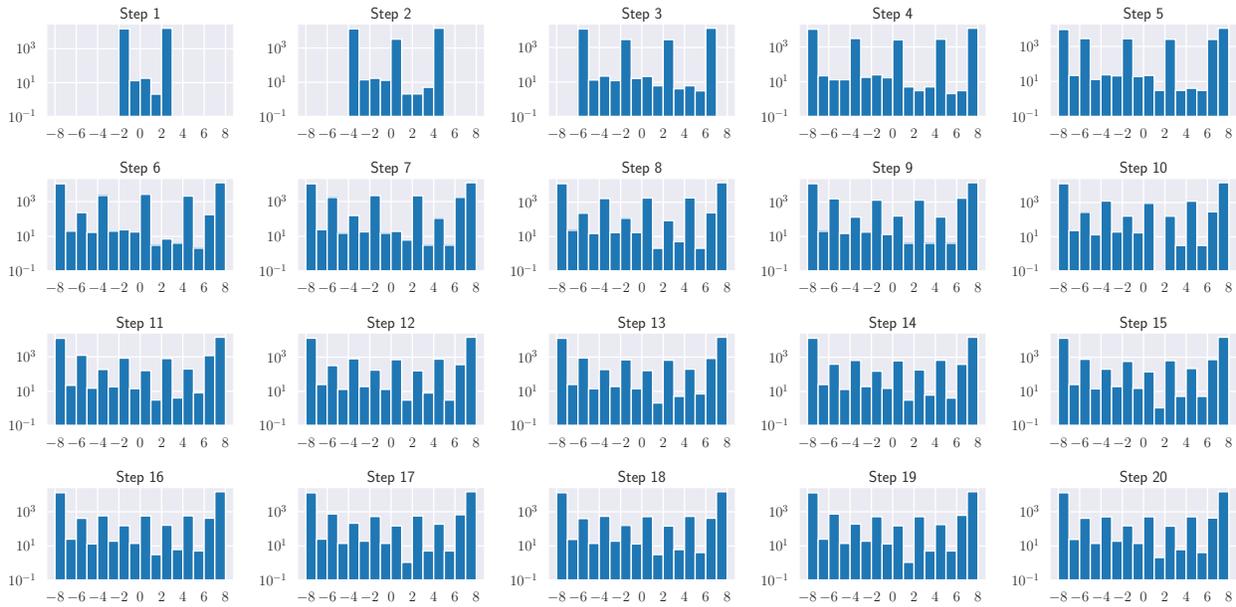}
    \caption{The figure shows the distribution over component values of PGD perturbation with the step size 2 over 20 attack steps, denoted as PGD-2-20. With the increasing attack steps, the PGD gradients get more components with bigger magnitude and gradually converges with bias.}
    \label{figure_app_pgd20_size2}
\end{figure*}

\begin{figure*}[tbp]
    \centering
    \includegraphics[width=0.95\textwidth]{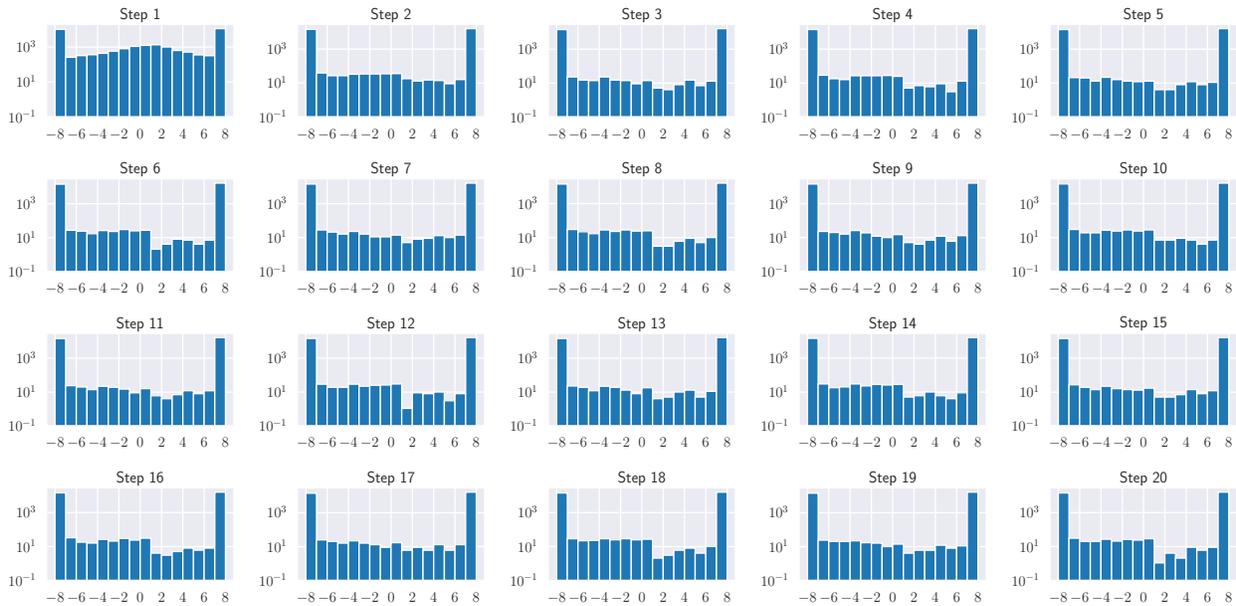}
    \caption{The figure shows the distribution over component values of SPGD perturbation with step size of $75,000,000$ over 20 attack steps, denoted as SPGD-75e6-20. With the increasing attack steps, the SPGD gradients get more components with large magnitude at the first step and it has local optimal components inside the boundary. Then the distribution quickly converges.}
    \label{figure_app_spgd20}
\end{figure*}

\begin{figure*}[tbp]
    \centering
    \includegraphics[width=0.95\textwidth]{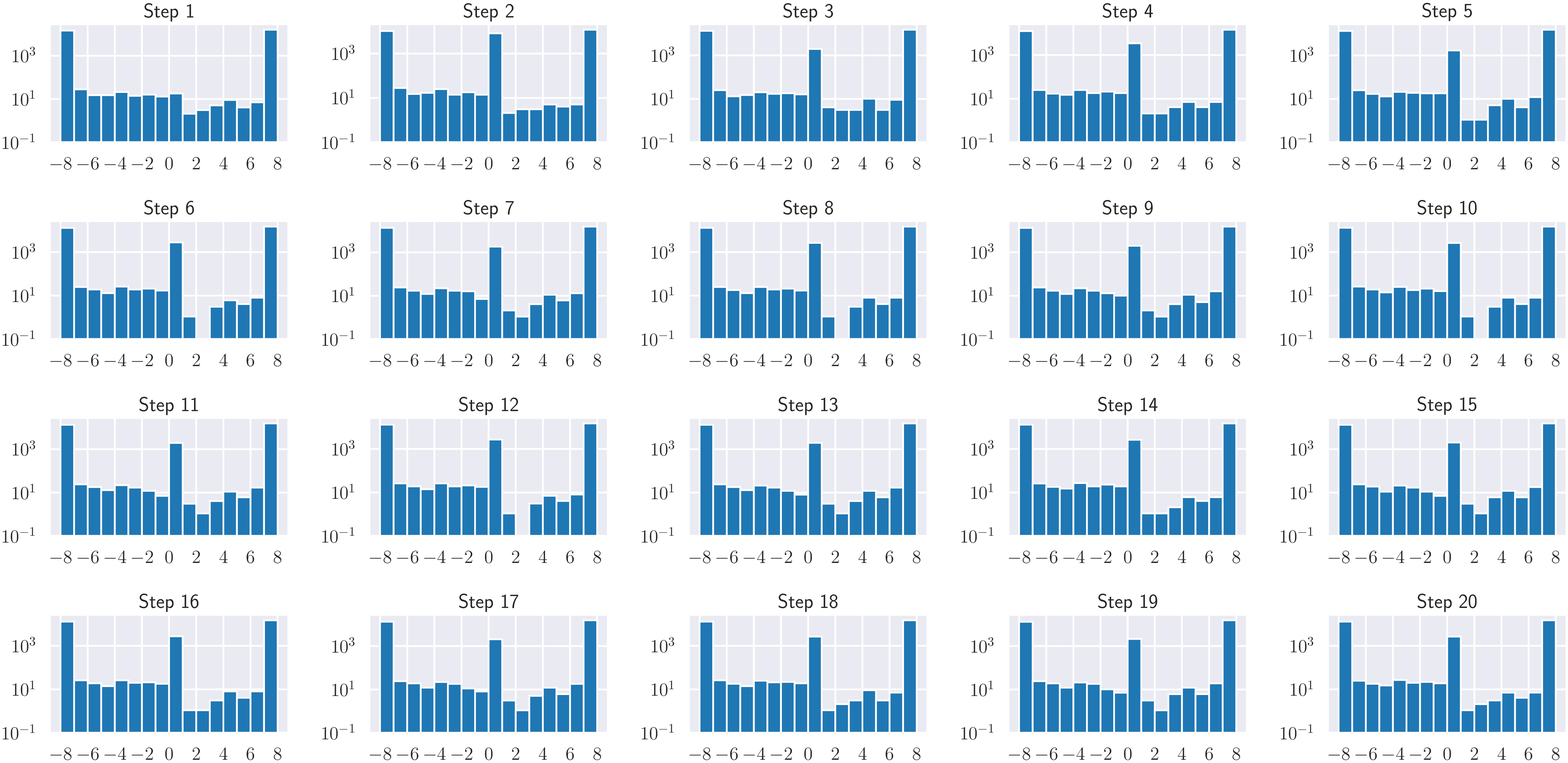}
    \caption{The figure shows the distribution over component values of PGD perturbation with step size of 8 over 20 attack steps, denoted as PGD-8-20. With the increasing attack steps, the PGD gradients gets some components with large magnitude at the first step, but its small components projected back from the outside of the boundary are not local optimal. Then the distribution gradually converges with bias. }
    \label{figure_app_pgd20_size8}
\end{figure*}

We calculated the component distribution plots of the adversarial perturbation of randomly picked 10 images from CIFAR10 dataset over 20 attack steps, and show them as \Cref{figure_app_pgd20_size2} for PGD with step size 2 denoted as PGD-2-20, \Cref{figure_app_spgd20} for SPGD with step size 75,000,000 denoted as SPGD-75e6-20 and \Cref{figure_app_pgd20_size8} denoted as PGD-8-20 for PGD with step size 8.

In \Cref{figure_app_pgd20_size2}, we show that, at the beginning, the spread of the distribution is restricted to the attack step size, and then the distribution gets wider toward the epsilon boundary with the increase of attack step.
In \Cref{figure_app_spgd20}, we show that, at the beginning, the spread of the distribution is wider and concentrated on the boundary with the local optimal small components in between, and then the components get more and more toward the epsilon boundary with the increase of attack step.
In \Cref{figure_app_pgd20_size8}, we show that, at the beginning, the spread of the distribution is also wider and concentrated on the boundary, but the small components are projected back by the clipping operation to the inside boundary.
They are not locally optimal, and then the distribution of the components does not have big change due to the restriction to the larger step size.


\section{MNIST Experiments}
\label{app:mnist}
We used the model architecture of two convolutional layers followed by two fully connected layers. Each convolutional layer is followed by  a $2 \times 2$ max-pooling layer.
In this case, the natural accuracy of all adversarially-trained models is over $98\%$.
We run 40 attack steps as our adversary during adversarial training, and use 100 steps of PGD or SPGD as our adversary for evaluation.
The step size we applied is 0.01 for PGD and 100 for SPGD, and we constrain both attacks to an $l_{\infty}$ norm ball of radius $\epsilon=0.3$.

\subsubsection{Adversarial Training on MNIST}
\begin{table}[ht]
    \centering
    \caption{The adversarial accuracy of the adversarially-trained models with SPGD or PGD on the MNIST dataset. We adversarially trained the models with $\epsilon=0.3$, using the SPGD attack and the PGD attack variously. We also reduce the number of attack steps used during adversarial training from 40 steps to 20 steps for both cases. We used either the SPGD or PGD attack to evaluate the adversarially-trained models and found that the SPGD-trained model retained better robustness when using 20 attack steps during training than the corresponding PGD-trained model.}
    \label{tab:security_curve_mnist}
    \begin{adjustbox}{width=\columnwidth}
        \begin{tabular}{lcccccccccc}
            \toprule
            \multicolumn{11}{c}{MNIST} \\
            \multicolumn{2}{r}{Attack Strength ($\epsilon$)} & 0 & 0.01 & 0.05 & 0.1 & 0.2 & 0.3 & 0.4 & 0.5 & 0.6\\
            \midrule
            Attack Method & Defense Model & \multicolumn{6}{l}{Adversarial Accuracy} \\
            \cmidrule(r){1-1} \cmidrule(lr){2-2} \cmidrule(l){3-11}
            PGD-0.01-100 & SPGD-100-40 & 98.73\% & 98.56\% & 98.10\% & 97.21\% & 94.74\% & 90.37\% &  1.06\% & 0.00\% & 0.00\% \\
            PGD-0.01-100 & PGD-0.01-40  & 98.42\% & 98.30\% & 97.56\% & 96.84\% & 94.34\% & 90.96\% & 20.95\% & 0.09\% & 0.00\% \\
            PGD-0.01-100 & SPGD-100-20 & 98.76\% & 98.57\% & 97.98\% & 97.09\% & 93.87\% & 87.88\% &  0.02\% & 0.00\% & 0.00\% \\
            PGD-0.01-100 & PGD-0.01-20  & 99.03\% & 98.85\% & 98.31\% & 97.12\% & 92.42\% & 80.33\% &  0.05\% & 0.00\% & 0.00\% \\ 
            \addlinespace[4pt]
            SPGD-100-100 & SPGD-100-40 & 98.73\% &  98.56\% & 98.14\% & 97.45\% & 95.78\% & 94.24\% & 23.49\% & 1.01\% & 0.20\% \\
            SPGD-100-100 & PGD-0.01-40  & 98.42\% &  98.28\% & 97.54\% & 96.87\% & 94.62\% & 91.57\% & 35.75\% & 3.00\% & 0.69\% \\
            SPGD-100-100 & SPGD-100-20 & 98.76\% &  98.58\% & 98.01\% & 97.35\% & 95.18\% & 93.08\% & 20.03\% & 1.23\% & 0.77\% \\
            SPGD-100-100 & PGD-0.01-20  & 99.03\% &  98.86\% & 98.34\% & 97.43\% & 94.09\% & 86.59\% & 19.95\% & 4.42\% & 0.86\% \\ 
            \bottomrule
        \end{tabular}
    \end{adjustbox}
\end{table}

As shown in \Cref{tab:security_curve_mnist}, the adversarial accuracy of the SPGD-100-40-trained model is competitive with the PGD-0.01-40-trained model within the threat model of $\epsilon=0.3$.
When the $\epsilon=0.3$, SPGD-100-40-trained performs the adversarial accuracy at 90.37\% close to the PGD-0.01-40-trained at 90.96\% under the attack of the PGD, and the SPGD-100-40-trained performs the adversarial accuracy at 94.24\% better than the PGD-0.01-40-trained at 91.57\% under the attack of the SPGD.

As shown in \Cref{tab:security_curve_mnist}, the adversarial accuracy of the SPGD-100-20-trained model has less performance drop from the SPGD-100-40-trained model, because the SPGD provides earlier convergence of attack than the PGD. 
When under the PGD attack with $\epsilon=0.3$, the SPGD-100-20-trained model performs at 87.88\% adversarial accuracy lower than the SPGD-100-40-trained model at 90.37\% by only 2.49\% points; however, the PGD-0.01-20-trained model performs at 80.33\% adversarial accuracy much lower than the PGD-0.01-40-trained model at 90.96\% by 10.63\% points.
When under the SPGD attack with $\epsilon=0.3$, the SPGD-100-20-trained model performs at 94.24\% adversarial accuracy lower than the SPGD-100-40-trained model at 93.08\% by only 1.16\%; however, the PGD-0.01-20-trained model performs at 86.59\% much lower than the PGD-0.01-40-trained model 91.57\% by 4.98\% points.
In both SPGD and PGD cases, the SPGD-trained model has less performance drop in fewer-step adversarial training. 
Whether being attacked by SPGD or PGD, the SPGD-trained model retains robustness much better than the PGD-trained model by using fewer attack steps during adversarial training. 
When under the PGD attack with $\epsilon=0.3$, the SPGD-100-20-trained model performs at 87.88\% adversarial accuracy higher than the PGD-0.01-20-trained model at 80.33\% by only 7.55\% points.
When under the SPGD attack with $\epsilon=0.3$, the SPGD-100-20-trained model performs at 93.08\% adversarial accuracy higher than the PGD-0.01-20-trained model at 86.59\% by 6.49\% points.
Whether being attacked by SPGD or PGD, the SPGD-trained model is more robust than the PGD-trained model in fewer-step adversarial training.

\end{document}